\pdfoutput=1

\documentclass[11pt]{article}

\usepackage[T1]{fontenc}

\usepackage[utf8]{inputenc}

\newcommand\camready[1]{}

\usepackage[hyperref]{acl}
\usepackage{times}
\usepackage{amsmath}
\usepackage{latexsym}
\usepackage[capitalize,noabbrev]{cleveref}

\usepackage{graphicx}
\usepackage{xcolor}
\usepackage{multirow}
\usepackage{makecell}
\usepackage{booktabs} 
\usepackage{subcaption}
\usepackage{siunitx}
\usepackage{xurl}  
\usepackage{enumitem}  
\usepackage{makecell}  
\usepackage{arydshln}

\usepackage{microtype}

\usepackage{scalerel,stackengine}
\stackMath
\newcommand\reallywidehat[1]{%
\savestack{\tmpbox}{\stretchto{%
  \scaleto{%
    \scalerel*[\widthof{\ensuremath{#1}}]{\kern-.6pt\bigwedge\kern-.6pt}%
    {\rule[-\textheight/2]{1ex}{\textheight}}
  }{\textheight}%
}{0.5ex}}%
\stackon[1pt]{#1}{\tmpbox}%
}
\title{What Do You Get When You Cross\\Beam Search with Nucleus Sampling?}

\author{Uri Shaham \qquad Omer Levy \\
The Blavatnik School of Computer Science\\
Tel Aviv University}

\date{}

\begin{document}
\maketitle

\begin{abstract}
We combine beam search with the probabilistic pruning technique of nucleus sampling to create two deterministic \textit{nucleus search} algorithms for natural language generation.
The first algorithm, $p$-exact search, locally prunes the next-token distribution and performs an exact search over the remaining space.
The second algorithm, dynamic beam search, shrinks and expands the beam size according to the entropy of the candidate's probability distribution.
Despite the probabilistic intuition behind nucleus search, experiments on machine translation and summarization benchmarks show that both algorithms reach the same performance levels as standard beam search.
\end{abstract}

\section{Introduction}
\label{sec:introduction}

The standard approach to natural language generation uses a search algorithm, guided by an autoregressive (conditional) language model, to search through the space of possible strings.
Since this search space is immense, \camready{various }pruning techniques have been introduced to facilitate tractable text generation.
Beam search \cite{beamsearch} is a deterministic algorithm that prunes the search space according to the relative \textit{rank} of each prefix, keeping only the top $b$ prefixes at every step. 
Although rank-based pruning has no probabilistic justification -- it is mainly motivated by its ability to limit memory consumption -- beam search is an effective approach for \camready{conditional text }generation tasks\camready{,} such as machine translation and summarization.
Nucleus sampling \cite{Holtzman2020The}, on the other hand, is a stochastic algorithm, which prunes the bottom \textit{percentile} of the model's next-token distribution, thus eliminating bad candidates while retaining some degree of randomness, which is important for free-form \camready{text }generation.
What if we were to replace beam search's rank-based pruning mechanism (top $k$) with the probabilistic mechanism of nucleus sampling (top $p$)?

We experiment with two variants of this hypothetical \textit{nucleus search}.
The first algorithm, \textit{$p$-exact search}, locally prunes the search space by retaining only the top $p$ of every next-token distribution that the underlying language model produces.
It then performs an exact search over the remaining space, guaranteeing the most probable sequence under the local pruning assumption. The second algorithm, \textit{dynamic beam search}, selects the top $p$ \textit{beams} at each step, according to their normalized probabilities (rather than top $k$, by rank).
This method can \camready{effectively} shrink or enhance the number of beams to match the current step's low or high entropy, respectively.

We evaluate both algorithms on three \camready{different }conditional generation benchmarks: subword-level translation (WMT'14 EN-FR), character-level translation (IWSLT'14 DE-EN), and summarization (XSUM\camready{ with BART pretraining}).
While we observe that both nucleus search algorithms produce competitive results with standard beam search, we do not find any empirical advantage to our probabilistically-motivated approach. 

We further analyze the algorithms by isolating the impact of dynamically expanding or shrinking the number of candidates. 
Experiments show that expanding the beam, even when entropy is high, tends to decrease performance.
Pruning candidates, on the other hand, appears to have no adverse effects, and may even have a marginal positive effect in certain cases, which possibly cancels out with the negative effects of beam expansion.

\section{Background}
\label{sec:background}

Natural language generation can be defined as a search problem in the space of possible sequences over a token vocabulary $V$, where the goal is to find an optimal sequence $Y = (y_1, ..., y_n) \in V^*$ according to some cost function.
Typical search algorithms explore this infinite space via sequence prefixes, starting with the empty sequence, and\camready{ incrementally} appending one potential token $y_t$ at a time.
Search terminates by returning a sequence (or a sequences set) that ends with a special token that indicates the end of the sequence (\texttt{EOS}).

The cost function is based on an underlying language model that, given a prefix $Y_{<t}$, induces a probability distribution over $V$, which we denote $P(y_t|Y_{<t})$.\footnote{The underlying model is often a \textit{conditional} language model $P(y_t|Y_{<t}, X)$, which takes an additional sequence $X$ as part of its input. For brevity, we omit $X$ from our notation.}
The probability of a sequence (or prefix) $Y$ is computed as the product of its tokens probabilities:
\begin{align}
P(Y) = \prod_t P(y_t | Y_{<t})
\label{eqn:1}
\end{align}
In practice, it is common to use the negative log probability instead:
\begin{align}
- \log P(Y) = \sum_t - \log P(y_t | Y_{<t})
\label{eqn:2}
\end{align}
This defines a monotonic additive cost function, where appending each token $y_t$ adds a positive cost $- \log P(y_t | Y_{<t})$ to the total cost of the sequence.

\subsection{Beam Search}

In many natural language generation tasks, \textit{beam search} \cite{beamsearch} is the algorithm of choice.
It extends the simple greedy algorithm by considering $k$ possible prefixes $\{ Y_{\leq t}^i \}_{i=1}^{k}$ at each timestep.
The beam size $k$ is constant throughout the search, guaranteeing a limit on memory consumption.

At every step $t$, beam search ranks all the possible single-token extensions of the current $k$ prefixes, and then keeps only the best $k$ extensions according to their total cost (Equation \ref{eqn:2}).
Once a prefix is appended with \texttt{EOS}, it is considered a complete sequence, and remains fixed as long as its cost is among the best $k$ prefixes; if $k$ (or more) better prefixes are found, it is discarded.
The algorithm terminates when either the final token of all top $k$ sequences is \texttt{EOS}, or when $t$ exceeds the predefined maximum number of steps.
In both cases, it returns all sequences in the beam that end with \texttt{EOS}.\footnote{Typically, the system selects the top sequence in the set, or chooses an alternative sequence via some reranking criterion.}

Assuming the \camready{underlying} models are tuned, results should improve as the beam size $k$ increases.
However, this assumption does \textit{not} hold for contemporary models;
in practice, text quality deteriorates when using large values of $k$ \cite{koehn-knowles-2017-six}.
Furthermore, decoding with exact search \cite{dijkstra1959note} reveals that translation models often rank the empty string as the most probable sequence \cite{stahlberg-byrne-2019-nmt}.
Perhaps unintentionally, searching with small beam sizes mitigates this flaw.\footnote{a.k.a. the ``blessing'' of beam search \cite{meister-etal-2020-beam}.}

\subsection{Nucleus Sampling}
\label{sec:nucleus}

Deterministic search algorithms, such as beam search, try to generate the most probable sequence.
This is a desirable property when we have many constraints regarding the target output, as in translation or question answering.
However, tasks that require more creativity and diversity in language may benefit from \textit{stochastic} algorithms.

\citet{Holtzman2020The} show that sampling \camready{directly }from a language model's raw distribution $P$ produces degenerate text, and instead, suggest to sample only from the \textit{nucleus}, $S_p$: the smallest set of tokens whose sum of probabilities is larger than some hyperparameter $p$.
Specifically, nucleus sampling prunes $P$ by assigning zero probability to every token outside of $S_p$,
and \camready{then }renormalizes the probabilities to get a new distribution $P_p$:
\begin{align*}
P_p(y | Y_{<t}) = \begin{cases}
\frac{P(y | Y_{<t})}{\sum_{y^\prime \in S_p} P(y^\prime | Y_{<t})} & y \in S_p \\
0 & y \notin S_p
\end{cases}
\end{align*}
Here, we refer to this mechanism as \textit{tail pruning}.
Sampling from $P_p$ results in less degenerate and more human-like text than both full-distribution sampling and top-$k$ sampling \cite{fan-etal-2018-hierarchical}, which do not account for the distribution's entropy.



\section{Nucleus Search} \label{sec:nucleus_search}
We combine beam search with tail pruning, producing two variants of \textit{nucleus search}: \textit{p-exact search} and \textit{dynamic beam search}.

\subsection{$p$-Exact Search}
\label{sec:p_exact}

\citet{stahlberg-byrne-2019-nmt} show that exact search \cite{dijkstra1959note} often produces extremely short and even empty sequences because the underlying \camready{language} model assigns a non-zero probability to the \texttt{EOS} token at each step.
We use tail pruning (Section~\ref{sec:nucleus}) to round all near-zero probabilities (whether belonging to \texttt{EOS} or any other token) to\camready{ an absolute} zero.
We apply exact search over the pruned space, guaranteeing the most probable sequence that contains only top-$p$ tokens at each step.

Given a hyperparameter $p$, we apply tail pruning to the model's predicted token distribution $P(y_t | Y_{<t})$.
The pruned distribution $P_p (y_t | Y_{<t})$ assigns zero probability to all tokens in the bottom $1-p$ of the original distribution, and remonrmalized probabilities for the rest.
\camready{For example, if the model's distribution over the first token assigns $P(y_1 = \text{George}) = 0.567$, and the hyperparameter $p=0.5$, then the renormalized distribution $P_p$ will assign all its probability mass to the token \textit{George}.
Conversely, if the model predicts $P(y_1 = \text{George}) = 0.0001$, and this event is not in the top $p$ of the distribution, then the new distribution will assign $P_p(y_1 = \text{George}) = 0$ and effectively prune all sequences beginning with the token \textit{George} from being generated.}
This procedure prunes the \texttt{EOS} token when it is unlikely, preventing empty sequences and reducing the brevity bias\camready{ in general}.

\subsection{Dynamic Beam Search}

Beam search keeps a fixed number ($k$) of prefixes according to their \textit{rank}\camready{, regardless of their probability scores}.
When entropy is high, the difference between the $k$-th most probable prefix and the one ranked $k+1$ might be minuscule, and we may want the search algorithm to consider such candidate prefixes as well.
Conversely, when entropy is low\camready{ (which is the case for most timesteps)}, the best prefix dominates the alternatives, making them redundant.

Dynamic beam search provides a mechanism for increasing the beam size when entropy is high, and pruning the number of prefixes when entropy is low.
Let $k_t$ be the number of viable prefixes at step $t$.
The model predicts the next-token distribution for each prefix, creating $k_t \cdot |V|$ candidates.
Each candidate $Y^i$ is scored according to its \textit{cumulative} probability $P(Y^i)$ (Equation \ref{eqn:1}).
To determine the beam size, we first normalize the probability scores within the set of candidates, and then apply tail pruning on the normalized probability:
\begin{align*}
\hat{P}(Y^i) = \frac{P(Y^i)}{\sum_{j=1}^{k_t \cdot |V|} P(Y^j)}
\end{align*}
As in $p$-exact search (Section~\ref{sec:p_exact}), we use a hyperparameter $p$ to determine the nucleus of $\hat{P}$, and thus the size of the next step's beam $k_{t+1}$.
The normalized probability $\hat{P}(Y^i) $ is only used to compute the dynamic beam; we keep the original probability $P(Y^i)$ as each prefix's cumulative score.

\section{Experiments}

We compare our search algorithms to beam search on a variety of tasks,\camready{\footnote{We do not compare to stochastic algorithms such as nucleus sampling \cite{Holtzman2020The}, since those are more suited for free-form language generation, while we focus on conditional text generation.}}
and use the same model across all settings, for each task.

\subsection{Tasks}

\paragraph{Machine Translation}
We evaluate on the WMT'14 EN-FR dataset \cite{ws-2014-statistical}, using the model of \citet{ott-etal-2018-scaling}, a large Transformer  \cite{NIPS2017_3f5ee243} with 6 encoder and decoder layers, trained on 36M bilingual sentences\camready{.
The model uses BPE subword tokenization, with a joint vocabulary of 44k types.}, tokenized with BPE.
We evaluate the generated sequences using SacreBLEU \cite{post-2018-call}, case-sensitive, with the 13a tokenizer.

\paragraph{Character-Level Machine Translation}
We train a character-level model on the IWSLT'14 DE-EN dataset \cite{cettolo2014report}, which contains approximately 172k bilingual sentences in its training set.
We use the recommended settings in Fairseq \cite{ott-etal-2019-fairseq} for a 6-layer encoder-decoder transformer.
As with the subword-level dataset, performance is measured via SacreBLEU.

\paragraph{Summarization}
We evaluate on the XSUM dataset \cite{narayan-etal-2018-dont}.
To alleviate memory issues and improve data quality, we remove examples where the source document is longer than 800 tokens (1,663 examples), or when the target \camready{summarization} is longer than one quarter of the source document (698 examples).
Our cleaned version of the XSUM test set contains 8,972 document-summarization pairs.
We use the large fine-tuned BART model \cite{lewis-etal-2020-bart}, and compute
ROUGE-L \cite{lin-hovy-2003-automatic} via compare-mt \cite{neubig-etal-2019-compare}.

\subsection{Implementation}
\label{par:impl}

Although both nucleus search algorithms can theoretically consume an unbounded amount of memory, our implementation caps the number of candidate prefixes by a large constant: 320 for WMT'14 and XSUM, and 160 for character-level translation.


We explore $p$ in increments of 0.1 for both nucleus search algorithms.
For beam search, we experiment with all beam sizes from 1 to 5, as well as exponentially increasing beam sizes from 5 to 320.
To present a complete picture of the algorithms' behaviors, we report results for all hyperparameter settings, rather than selecting the best configuration according to the validation set.
This experiment design limits our ability to claim the superiority of one algorithm over another, but as we show in Section \ref{sec:results}, the performance differences are so small that no such claim will be made.

\begin{table}[t]
\small
\centering
\begin{tabular}{@{}lcccc@{}}
\toprule
\multirow{2}{*}{\textbf{Search}} & \textbf{Hyper-} &  \multirow{2}{*}{\textbf{WMT'14}} & \textbf{IWSLT'14} & \multirow{3}{*}{\textbf{XSUM}} \\
\multirow{2}{*}{\textbf{Algo}} & \textbf{param} & \multirow{2}{*}{\textbf{EN-FR}} & \textbf{DE-EN} & \\
 & \textbf{(}$k$ \textbf{or} $p$\textbf{)} &  & \textbf{(Char)} & \\
\midrule
\multirow{11}{*}{Beam} & 1  &  40.3 & 33.3 & 35.5 \\ 
& 2  &  \underline{40.7} & \underline{33.6} & 36.2 \\ 
& 3  &  \textbf{40.8} & \underline{33.6} & \underline{36.4} \\ 
& 4  &  \textbf{40.8} & \underline{33.6} & \underline{36.5} \\ 
& 5  &  \underline{40.6} & \underline{33.5} & \underline{36.5} \\ 
& 10  &  40.5 & \underline{33.5} & \textbf{36.6} \\ 
& 20  &  40.2 & 33.1 & \underline{36.4} \\ 
& 40  &  39.6 & 27.4 & 36.1 \\ 
& 80  &  38.7 & 18.1 & 35.7 \\ 
& 160  &  32.2 & 5.3 & 34.3 \\ 
& 320  &  11.8 & 5.3 & 28.1 \\ 

\midrule
\multirow{9}{*}{$p$-Exact} & 0.1  &  40.3 & 33.3 & 35.5 \\ 
& 0.2  &  40.3 & 33.3 & 35.7 \\ 
& 0.3  &  40.5 & 33.3 & 36.1 \\ 
& 0.4  &  40.5 & 33.4 & \underline{36.5} \\ 
& 0.5  &  \underline{40.6} & \underline{33.5} & \textbf{36.6} \\ 
& 0.6  &  \underline{40.6} & \underline{33.5} & \textbf{36.6} \\ 
& 0.7  &  40.2 & \underline{33.6} & 36.3 \\ 
& 0.8  &  39.2 & \underline{33.6} & 35.9 \\ 
& 0.9  &  27.8 & 33.2 & 33.1 \\ 

\midrule
\multirow{9}{*}{Dynamic} & 0.1  &  40.2 & 33.3 & 35.5 \\ 
& 0.2  &  40.3 & 33.3 & 35.6 \\ 
& 0.3  &  40.5 & 33.4 & 36.0 \\ 
& 0.4  &  \underline{40.6} & 33.4 & 36.2 \\ 
& 0.5  &  \underline{40.6} & 33.4 & \underline{36.5} \\ 
& 0.6  &  \underline{40.6} & \textbf{33.7} & \underline{36.5} \\ 
& 0.7  &  40.0 & \textbf{33.7} & 36.0 \\ 
& 0.8  &  38.9 & \underline{33.6} & 35.4 \\ 
& 0.9  &  18.1 & 33.1 & 31.5 \\ 

\bottomrule
\end{tabular}
\caption{Scores of different algorithms and settings on various generation tasks. \textbf{Bold} numbers indicate the highest result on the task, and \underline{underlined} numbers indicate that the result is within 0.2 points of the top score.}
\label{tab:unnorm}
\vspace{-5pt}
\end{table}


\section{Results}
\label{sec:results}

\paragraph{Main Result}
Table \ref{tab:unnorm} shows the performance of each search algorithm across the different tasks.\footnote{This table shows performance without reranking (length normalization), to study the core algorithm. Appendix~\ref{appx:reranking} contains the results with reranking, showing similar trends.}
In line with previously reported trends \cite{koehn-knowles-2017-six}, we observe that increasing the beam size beyond $k=10$ can severely degrade performance.
On the other hand, the probabilistic search algorithms appear to be more stable, with most hyperparameter settings achieving relatively high performance metrics until $p = 0.9$, where substantial performance degradation is evident.

Despite their increased stability, there appears to be no significant advantage to either $p$-exact search or dynamic beam search over the original beam search.
In fact, the performance differences between the best settings of each algorithm are always under 0.2 BLEU/ROUGE, and often zero.
We find this trend counter-intuitive, since we originally assumed that expanding and trimming the beam based on entropy would benefit language generation. We further test these assumptions individually.

\paragraph{Expanded Beams}
We compare the performance of static beam search ($k=5$) and dynamic beam search ($p=0.6$) on two subsets of the translation task's test set:\footnote{We select $p=0.6$ since it is the maximal value that achieved the top score on the WMT'14 EN-FR benchmark.} (1) examples where dynamic beam search always selects from its top 5 prefixes, and (2) the complement, where every generated output contains at least one prefix that was ranked 6th or worse.
Table \ref{tab:unnorm_maxrank} shows that in those cases where dynamic beam search actually uses the expanded beam, i.e. it chooses prefixes that rank lower than 5, it performs \textit{worse} than static top-5 beam search by 0.7 BLEU.
This subset accounts for only 13\% of examples -- which are probably harder for the model, given the 10-point difference in BLEU -- while the majority 87\% of cases are always composed from the top 5 (or less) prefixes.

\begin{table}[t]
\small
\centering
\begin{tabular}{@{}llrr@{}}
\toprule
\multicolumn{2}{@{}l}{\textbf{Search Algorithm}}  &  $\max(i) \le 5$ & $\max(i) > 5$ \\
\midrule
Beam & $k=5$  &  42.2  & \textbf{32.9} \\ 
Dynamic Beam & $p=0.6$  & \textbf{42.3} & 32.2 \\ 
\midrule
\textit{\#Examples} &   &  \textit{2618} & \textit{385} \\ 
\bottomrule
\end{tabular}
\caption{Performance on two subsets of WMT'14 EN-FR: (1) examples where dynamic beam search only selects prefixes from the top-5 options ($\max(i) \le 5$), and (2) examples where the output of dynamic beam search contains at least one prefix that ranked 6 or worse ($\max(i) > 5$).}
\label{tab:unnorm_maxrank}
\vspace{-10pt}
\end{table}

\paragraph{Trimmed Beams}
We isolate the effect of probabilistic trimming by applying a $k=5$ cap on the number of active beams, for both nucleus search variations.
Table \ref{tab:unnorm_5} shows that $p$-exact and dynamic beam trimming strategies have no negative effects, and may have a marginal positive effect.

\begin{table}[t]
\small
\centering
\begin{tabular}{@{}lcccc@{}}
\toprule
\multirow{2}{*}{\textbf{Search}} & \textbf{Hyper-} &  \multirow{2}{*}{\textbf{WMT'14}} & \textbf{IWSLT'14} & \multirow{3}{*}{\textbf{XSUM}} \\
\multirow{2}{*}{\textbf{Algo}} & \textbf{param} & \multirow{2}{*}{\textbf{EN-FR}} & \textbf{DE-EN} & \\
 & \textbf{(}$k$ \textbf{or} $p$\textbf{)} &  & \textbf{(Char)} & \\
\midrule
\multirow{5}{*}{Beam} 
& 1  &  40.3 & 33.3  & 35.5 \\ 
& 2  & 40.7 & \underline{33.6}  & 36.2 \\ 
& 3  &  \underline{40.8} & \underline{33.6} &  \underline{36.4} \\ 
& 4  &  \underline{40.8} & \underline{33.6} &  \underline{36.5} \\ 
& 5  &  40.6 & 33.5 &  \underline{36.5} \\ 
\midrule
\multirow{8}{*}{$p$-Exact} & 0.1  &  40.3 & 33.3 & 35.5 \\ 
\multirow{8}{*}{($k=5$)} & 0.2  &  40.3 & 33.3  & 35.7 \\ 
& 0.3  &  40.5 & 33.3 &  36.1 \\ 
& 0.4  &  40.6 & 33.4 &  \underline{36.4} \\ 
& 0.5  &  \underline{40.8} & 33.5 &  \textbf{36.6} \\ 
& 0.6  &  \textbf{41.0} & \underline{33.6} &  \textbf{36.6} \\ 
& 0.7  &  \underline{40.9} & \underline{33.7} & \textbf{36.6} \\ 
& 0.8  &  \underline{40.9} & \textbf{33.8} &  \underline{36.5} \\ 
& 0.9  &  \underline{40.8} & \textbf{33.8} &  \underline{36.5} \\
\midrule
\multirow{8}{*}{Dynamic} & 0.1  &  40.2 & 33.3  & 35.5 \\ 
\multirow{8}{*}{($k=5$)} & 0.2  &  40.3 & 33.3 & 35.6 \\ 
& 0.3  &  40.5 & 33.4 &  36.0 \\ 
& 0.4  &  40.6 & 33.4 &  36.2 \\ 
& 0.5  &  40.6 & 33.4 & \underline{36.4} \\ 
& 0.6  &  \underline{40.8} & \underline{33.7} & \underline{36.5} \\ 
& 0.7  &  40.7 & \underline{33.7} & \textbf{36.6} \\ 
& 0.8  &  40.7 & \underline{33.6} &  \textbf{36.6} \\ 
& 0.9  &  40.6 & 33.5 &  \underline{36.5} \\ 
\bottomrule
\end{tabular}
\caption{Scores of different algorithms and settings on various generation tasks, \textit{when limiting the beam size to a maximum of 5}. \textbf{Bold} numbers indicate the highest result on the task, and \underline{underlined} numbers indicate that the result is within 0.2 points of the top score.}
\label{tab:unnorm_5}
\vspace{-10pt}
\end{table}

\section{Related Work}
\label{sec:RelatedWork}

As a standard decoding strategy, there is a significant body of literature on beam search.
Recently, there has been more focus on the empty string problem \cite{stahlberg-byrne-2019-nmt}, and the fact that increasing the beam size beyond a small constant typically hurts performance.
\citet{meister-etal-2020-beam} show that beam search optimize for sequences that distribute information uniformly, and therefore, using small beam sizes allows it to overcome the empty string problem.
\citet{shi2020neural} train models with multiple different \texttt{EOS} tokens based on their positions, instead of a single universal \texttt{EOS} token. 
\citet{peters-martins-2021-smoothing} replace the softmax function with the sparse entmax transformation \cite{peters-etal-2019-sparse} that \textit{can} assign absolute zero probability to tokens.
This method has a similar effect to our $p$-exact search, but requires training the model with entmax, while our contribution only modifies the search algorithm.

 \citet{massarelli-etal-2020-decoding} also propose a combination of beam search and sampling methods, but with a different method and a different goal. They focus on free-form text generation, addressing two problems -- repetition and halucination -- by sampling the first few tokens, and then switching over to beam search. \citet{freitag-al-onaizan-2017-beam} explore how using a small fixed beam size, pruned further according to the relative or absolute distance from the top scored candidate, can increase decoding speed. In this work, we focus on the quality of the generated text, comparing the use of a fixed beam size to tail pruning, an established method that keeps candidates according to the nucleus of the distribution.

\section{Conclusion}

Language models predict a distribution over their vocabulary, yet beam search only utilizes the rank of different candidates, not their actual probability scores.
A natural assumption is that searching the space of prefixes with a constant number of options is not optimal. 
We hypothesize that using the probability scores to dynamically determine the number of candidates may benefit natural language generation.
We test our hypothesis by introducing two nucleus search algorithms, which incorporate probabilistic tail pruning \cite{Holtzman2020The} with beam search, but find that they perform on par with the baseline beam search algorithm when its beam size is restricted to a small constant.








\section*{Acknowledgements}
This work was supported by the Tel Aviv University Data Science Center, the Blavatnik Fund, the Alon Scholarship, and Intel Corporation.
We would like to thank Ari Holtzman, Jonathan Berant, Ori Yoran, Lior Vassertail, and Yuval Kirstain for their valuable feedback.

\bibliographystyle{acl_natbib}
\bibliography{anthology,custom}

\begin{thebibliography}{24}
\expandafter\ifx\csname natexlab\endcsname\relax\def\natexlab#1{#1}\fi

\bibitem[{Bojar et~al.(2014)Bojar, Buck, Federmann, Haddow, Koehn, Monz, Post,
  and Specia}]{ws-2014-statistical}
Ond{\v{r}}ej Bojar, Christian Buck, Christian Federmann, Barry Haddow, Philipp
  Koehn, Christof Monz, Matt Post, and Lucia Specia, editors. 2014.
\newblock \href {https://doi.org/10.3115/v1/W14-33} {\emph{Proceedings of the
  Ninth Workshop on Statistical Machine Translation}}. Association for
  Computational Linguistics, Baltimore, Maryland, USA.

\bibitem[{Cettolo et~al.(2014)Cettolo, Niehues, St{\"u}ker, Bentivogli, and
  Federico}]{cettolo2014report}
Mauro Cettolo, Jan Niehues, Sebastian St{\"u}ker, Luisa Bentivogli, and
  Marcello Federico. 2014.
\newblock Report on the 11th iwslt evaluation campaign, iwslt 2014.
\newblock In \emph{Proceedings of the International Workshop on Spoken Language
  Translation, Hanoi, Vietnam}, volume~57.

\bibitem[{Dijkstra(1959)}]{dijkstra1959note}
Edsger~W Dijkstra. 1959.
\newblock A note on two problems in connexion with graphs.
\newblock \emph{Numerische mathematik}, 1(1):269--271.

\bibitem[{Fan et~al.(2018)Fan, Lewis, and Dauphin}]{fan-etal-2018-hierarchical}
Angela Fan, Mike Lewis, and Yann Dauphin. 2018.
\newblock \href {https://doi.org/10.18653/v1/P18-1082} {Hierarchical neural
  story generation}.
\newblock In \emph{Proceedings of the 56th Annual Meeting of the Association
  for Computational Linguistics (Volume 1: Long Papers)}, pages 889--898,
  Melbourne, Australia. Association for Computational Linguistics.

\bibitem[{Freitag and Al-Onaizan(2017)}]{freitag-al-onaizan-2017-beam}
Markus Freitag and Yaser Al-Onaizan. 2017.
\newblock \href {https://doi.org/10.18653/v1/W17-3207} {Beam search strategies
  for neural machine translation}.
\newblock In \emph{Proceedings of the First Workshop on Neural Machine
  Translation}, pages 56--60, Vancouver. Association for Computational
  Linguistics.

\bibitem[{Holtzman et~al.(2020)Holtzman, Buys, Du, Forbes, and
  Choi}]{Holtzman2020The}
Ari Holtzman, Jan Buys, Li~Du, Maxwell Forbes, and Yejin Choi. 2020.
\newblock \href {https://openreview.net/forum?id=rygGQyrFvH} {The curious case
  of neural text degeneration}.
\newblock In \emph{International Conference on Learning Representations}.

\bibitem[{Jean et~al.(2015)Jean, Firat, Cho, Memisevic, and
  Bengio}]{jean-etal-2015-montreal}
S{\'e}bastien Jean, Orhan Firat, Kyunghyun Cho, Roland Memisevic, and Yoshua
  Bengio. 2015.
\newblock \href {https://doi.org/10.18653/v1/W15-3014} {{M}ontreal neural
  machine translation systems for {WMT}{'}15}.
\newblock In \emph{Proceedings of the Tenth Workshop on Statistical Machine
  Translation}, pages 134--140, Lisbon, Portugal. Association for Computational
  Linguistics.

\bibitem[{Koehn and Knowles(2017)}]{koehn-knowles-2017-six}
Philipp Koehn and Rebecca Knowles. 2017.
\newblock \href {https://doi.org/10.18653/v1/W17-3204} {Six challenges for
  neural machine translation}.
\newblock In \emph{Proceedings of the First Workshop on Neural Machine
  Translation}, pages 28--39, Vancouver. Association for Computational
  Linguistics.

\bibitem[{Lewis et~al.(2020)Lewis, Liu, Goyal, Ghazvininejad, Mohamed, Levy,
  Stoyanov, and Zettlemoyer}]{lewis-etal-2020-bart}
Mike Lewis, Yinhan Liu, Naman Goyal, Marjan Ghazvininejad, Abdelrahman Mohamed,
  Omer Levy, Veselin Stoyanov, and Luke Zettlemoyer. 2020.
\newblock \href {https://doi.org/10.18653/v1/2020.acl-main.703} {{BART}:
  Denoising sequence-to-sequence pre-training for natural language generation,
  translation, and comprehension}.
\newblock In \emph{Proceedings of the 58th Annual Meeting of the Association
  for Computational Linguistics}, pages 7871--7880, Online. Association for
  Computational Linguistics.

\bibitem[{Lin and Hovy(2003)}]{lin-hovy-2003-automatic}
Chin-Yew Lin and Eduard Hovy. 2003.
\newblock \href {https://www.aclweb.org/anthology/N03-1020} {Automatic
  evaluation of summaries using n-gram co-occurrence statistics}.
\newblock In \emph{Proceedings of the 2003 Human Language Technology Conference
  of the North {A}merican Chapter of the Association for Computational
  Linguistics}, pages 150--157.

\bibitem[{Massarelli et~al.(2020)Massarelli, Petroni, Piktus, Ott,
  Rockt{\"a}schel, Plachouras, Silvestri, and
  Riedel}]{massarelli-etal-2020-decoding}
Luca Massarelli, Fabio Petroni, Aleksandra Piktus, Myle Ott, Tim
  Rockt{\"a}schel, Vassilis Plachouras, Fabrizio Silvestri, and Sebastian
  Riedel. 2020.
\newblock \href {https://doi.org/10.18653/v1/2020.findings-emnlp.22} {How
  decoding strategies affect the verifiability of generated text}.
\newblock In \emph{Findings of the Association for Computational Linguistics:
  EMNLP 2020}, pages 223--235, Online. Association for Computational
  Linguistics.

\bibitem[{Meister et~al.(2020)Meister, Cotterell, and
  Vieira}]{meister-etal-2020-beam}
Clara Meister, Ryan Cotterell, and Tim Vieira. 2020.
\newblock \href {https://doi.org/10.18653/v1/2020.emnlp-main.170} {If beam
  search is the answer, what was the question?}
\newblock In \emph{Proceedings of the 2020 Conference on Empirical Methods in
  Natural Language Processing (EMNLP)}, pages 2173--2185, Online. Association
  for Computational Linguistics.

\bibitem[{Murray and Chiang(2018)}]{murray-chiang-2018-correcting}
Kenton Murray and David Chiang. 2018.
\newblock \href {https://doi.org/10.18653/v1/W18-6322} {Correcting length bias
  in neural machine translation}.
\newblock In \emph{Proceedings of the Third Conference on Machine Translation:
  Research Papers}, pages 212--223, Brussels, Belgium. Association for
  Computational Linguistics.

\bibitem[{Narayan et~al.(2018)Narayan, Cohen, and
  Lapata}]{narayan-etal-2018-dont}
Shashi Narayan, Shay~B. Cohen, and Mirella Lapata. 2018.
\newblock \href {https://doi.org/10.18653/v1/D18-1206} {Don{'}t give me the
  details, just the summary! topic-aware convolutional neural networks for
  extreme summarization}.
\newblock In \emph{Proceedings of the 2018 Conference on Empirical Methods in
  Natural Language Processing}, pages 1797--1807, Brussels, Belgium.
  Association for Computational Linguistics.

\bibitem[{Neubig et~al.(2019)Neubig, Dou, Hu, Michel, Pruthi, and
  Wang}]{neubig-etal-2019-compare}
Graham Neubig, Zi-Yi Dou, Junjie Hu, Paul Michel, Danish Pruthi, and Xinyi
  Wang. 2019.
\newblock \href {https://doi.org/10.18653/v1/N19-4007} {compare-mt: A tool for
  holistic comparison of language generation systems}.
\newblock In \emph{Proceedings of the 2019 Conference of the North {A}merican
  Chapter of the Association for Computational Linguistics (Demonstrations)},
  pages 35--41, Minneapolis, Minnesota. Association for Computational
  Linguistics.

\bibitem[{Ott et~al.(2019)Ott, Edunov, Baevski, Fan, Gross, Ng, Grangier, and
  Auli}]{ott-etal-2019-fairseq}
Myle Ott, Sergey Edunov, Alexei Baevski, Angela Fan, Sam Gross, Nathan Ng,
  David Grangier, and Michael Auli. 2019.
\newblock \href {https://doi.org/10.18653/v1/N19-4009} {fairseq: A fast,
  extensible toolkit for sequence modeling}.
\newblock In \emph{Proceedings of the 2019 Conference of the North {A}merican
  Chapter of the Association for Computational Linguistics (Demonstrations)},
  pages 48--53, Minneapolis, Minnesota. Association for Computational
  Linguistics.

\bibitem[{Ott et~al.(2018)Ott, Edunov, Grangier, and
  Auli}]{ott-etal-2018-scaling}
Myle Ott, Sergey Edunov, David Grangier, and Michael Auli. 2018.
\newblock \href {https://doi.org/10.18653/v1/W18-6301} {Scaling neural machine
  translation}.
\newblock In \emph{Proceedings of the Third Conference on Machine Translation:
  Research Papers}, pages 1--9, Brussels, Belgium. Association for
  Computational Linguistics.

\bibitem[{Peters and Martins(2021)}]{peters-martins-2021-smoothing}
Ben Peters and Andr{\'e} F.~T. Martins. 2021.
\newblock \href {https://doi.org/10.18653/v1/2021.naacl-main.210} {Smoothing
  and shrinking the sparse {S}eq2{S}eq search space}.
\newblock In \emph{Proceedings of the 2021 Conference of the North American
  Chapter of the Association for Computational Linguistics: Human Language
  Technologies}, pages 2642--2654, Online. Association for Computational
  Linguistics.

\bibitem[{Peters et~al.(2019)Peters, Niculae, and
  Martins}]{peters-etal-2019-sparse}
Ben Peters, Vlad Niculae, and Andr{\'e} F.~T. Martins. 2019.
\newblock \href {https://doi.org/10.18653/v1/P19-1146} {Sparse
  sequence-to-sequence models}.
\newblock In \emph{Proceedings of the 57th Annual Meeting of the Association
  for Computational Linguistics}, pages 1504--1519, Florence, Italy.
  Association for Computational Linguistics.

\bibitem[{Post(2018)}]{post-2018-call}
Matt Post. 2018.
\newblock \href {https://doi.org/10.18653/v1/W18-6319} {A call for clarity in
  reporting {BLEU} scores}.
\newblock In \emph{Proceedings of the Third Conference on Machine Translation:
  Research Papers}, pages 186--191, Brussels, Belgium. Association for
  Computational Linguistics.

\bibitem[{Reddy(1977)}]{beamsearch}
D.~Raj Reddy. 1977.
\newblock \href {https://doi.org/10.1184/R1/6609821.v1} {Speech understanding
  systems: A summary of results of the five-year research effort at
  carnegie-mellon university}.

\bibitem[{Shi et~al.(2020)Shi, Xiao, and Knight}]{shi2020neural}
Xing Shi, Yijun Xiao, and Kevin Knight. 2020.
\newblock \href {http://arxiv.org/abs/2012.13454} {Why neural machine
  translation prefers empty outputs}.

\bibitem[{Stahlberg and Byrne(2019)}]{stahlberg-byrne-2019-nmt}
Felix Stahlberg and Bill Byrne. 2019.
\newblock \href {https://doi.org/10.18653/v1/D19-1331} {On {NMT} search errors
  and model errors: Cat got your tongue?}
\newblock In \emph{Proceedings of the 2019 Conference on Empirical Methods in
  Natural Language Processing and the 9th International Joint Conference on
  Natural Language Processing (EMNLP-IJCNLP)}, pages 3356--3362, Hong Kong,
  China. Association for Computational Linguistics.

\bibitem[{Vaswani et~al.(2017)Vaswani, Shazeer, Parmar, Uszkoreit, Jones,
  Gomez, Kaiser, and Polosukhin}]{NIPS2017_3f5ee243}
Ashish Vaswani, Noam Shazeer, Niki Parmar, Jakob Uszkoreit, Llion Jones,
  Aidan~N Gomez, \L~ukasz Kaiser, and Illia Polosukhin. 2017.
\newblock \href
  {https://proceedings.neurips.cc/paper/2017/file/3f5ee243547dee91fbd053c1c4a845aa-Paper.pdf}
  {Attention is all you need}.
\newblock In \emph{Advances in Neural Information Processing Systems},
  volume~30. Curran Associates, Inc.

\end{thebibliography}

\appendix
\newpage
~~~
\newpage
\section{Results with Reranking}
\label{appx:reranking}

When presenting our main results (Section~\ref{sec:results}), we follow related work \cite{peters-martins-2021-smoothing} and focus on the outputs generated using the algorithms themselves, without reranking.
For completeness, we also present the results of applying length normalization \cite{jean-etal-2015-montreal,murray-chiang-2018-correcting}, i.e. reranking the set of sequences produced by beam search according to their average log-probability, rather than their cumulative log-probability: 
\begin{align*}
\text{score}(Y) = \frac{1}{n} \sum_{t=1}^{n} - \log P(y_t | Y_{<t})
\end{align*}
Table \ref{tab:norm} shows that length normalization improves stability, and slightly increases performance overall.
However, it does \textit{not} increase the performance gap between the different algorithms, with respect to the results in Section~\ref{sec:results} (without reranking);
all three variants produce text that scores within 0.2 BLEU/ROUGE from the best performing setting in every task.

\begin{table}[t!]
\small
\centering
\begin{tabular}{@{}lcccc@{}}
\toprule
\multirow{2}{*}{\textbf{Search}} & \textbf{Hyper-} &  \multirow{2}{*}{\textbf{WMT'14}} & \textbf{IWSLT'14} & \multirow{3}{*}{\textbf{XSUM}} \\
\multirow{2}{*}{\textbf{Algo}} & \textbf{param} & \multirow{2}{*}{\textbf{EN-FR}} & \textbf{DE-EN} & \\
 & \textbf{(}$k$ \textbf{or} $p$\textbf{)} &  & \textbf{(Char)} & \\
\midrule
\multirow{11}{*}{Beam} & 1  &  40.3 & 33.3 & 35.5 \\ 
& 2  &  40.8 & 33.8 & 36.3 \\ 
& 3  &  \textbf{41.1} & \underline{34.0}& \underline{36.4} \\ 
& 4  &  \textbf{41.1} & \underline{34.1} & \underline{36.5} \\ 
& 5  &  \underline{41.0} & \underline{34.1} & \textbf{36.6} \\ 
& 10  &  \underline{41.0} & \textbf{34.2} & \textbf{36.6} \\ 
& 20  &  \underline{41.0} & \textbf{34.2} & \underline{36.5} \\ 
& 40  &  40.6 & \textbf{34.2} & \underline{36.4} \\ 
& 80  &  40.1 & \textbf{34.2} & 36.3 \\ 
& 160  &  39.4 & \textbf{34.2} & 36.2 \\ 
& 320  &  38.3 & \textbf{34.2} & 36.2 \\ 
\midrule
\multirow{9}{*}{$p$-Exact} & 0.1  &  40.3 & 33.3 & 35.5 \\ 
& 0.2  &  40.3 & 33.3 & 35.6 \\ 
& 0.3  &  40.5 & 33.4 & 36.0 \\ 
& 0.4  &  40.7 & 33.4 & 36.2 \\ 
& 0.5  &  \underline{41.0} & 33.6 & \underline{36.4} \\ 
& 0.6  &  \textbf{41.1} & 33.7 & 36.3 \\ 
& 0.7  &  \underline{41.0} & \underline{34.0} & 36.3 \\ 
& 0.8  &  40.3 & \underline{34.1} & 36.2 \\ 
& 0.9  &  38.8 & \underline{34.1} & 36.1 \\
\midrule
\multirow{9}{*}{Dynamic} & 0.1  &  40.2 & 33.3 & 35.5 \\ 
& 0.2  &  40.3 & 33.3 & 35.6 \\ 
& 0.3  &  40.5 & 33.4 & 36.0 \\ 
& 0.4  &  40.6 & 33.4 & 36.2 \\ 
& 0.5  &  40.8 & 33.4 & \underline{36.4} \\ 
& 0.6  &  \underline{41.0} & 33.8 & \underline{36.5} \\ 
& 0.7  &  \underline{41.0} & \underline{34.0} & 36.3 \\ 
& 0.8  &  40.6 & \underline{34.1} & 36.2 \\ 
& 0.9  &  38.6 & \textbf{34.2} & 36.2 \\ 
\bottomrule
\end{tabular}
\caption{The performance of different decoding algorithms and hyperparameter settings on various conditional generation tasks with \textit{length normalization (reranking)}. \textbf{Bold} numbers indicate the highest result on the task, and \underline{underlined} numbers indicate that the result is within 0.2 points of the top score. }
\label{tab:norm}
\end{table}

\end{document}